\documentclass[10pt,twocolumn,letterpaper]{article}

\usepackage{iccv}
\usepackage{times}
\usepackage{epsfig}
\usepackage{graphicx}
\usepackage{amsmath}
\usepackage{amssymb}

\usepackage{color}
\usepackage{times}
\usepackage{overpic}
\usepackage{bm}
\usepackage{tabu}
\usepackage{bbding}
\usepackage{multirow}
\usepackage[table]{xcolor}
\usepackage[pagebackref=true,breaklinks=true,letterpaper=true,colorlinks,bookmarks=false]{hyperref}

\iccvfinalcopy % *** Uncomment this line for the final submission

\def\iccvPaperID{3649} % *** Enter the ICCV Paper ID here

\ificcvfinal\fi

\begin{document}

%%%%%%%%% TITLE
\title{Human-centric Scene Understanding for 3D Large-scale Scenarios}

\author{\textbf{
Yiteng Xu$^{1,}$\thanks{Equal contribution. $\dagger$ Corresponding author. This work was supported by NSFC (No.62206173), Natural Science Foundation of Shanghai (No.22dz1201900), MoE Key Laboratory of Intelligent Perception and Human-Machine Collaboration (ShanghaiTech University), Shanghai Frontiers Science Center of Human-centered Artificial Intelligence (ShangHAI), Shanghai Engineering Research Center of Intelligent Vision and Imaging.}  ,  
Peishan Cong$^{1,}\footnote[1]{}$  , 
Yichen Yao$^{1,}\footnote[1]{}$  ,}\\
\textbf{
Runnan Chen$^{2}$, 
Yuenan Hou$^{3}$, 
Xinge Zhu$^{4}$, 
Xuming He$^{1}$, 
Jingyi Yu$^{1}$, 
Yuexin Ma$^{1,}$\footnote[2]{}}\\ 
$^{1}$ ShanghaiTech University
$^{2}$ The University of Hong Kong \\
$^{3}$ Shanghai AI Laboratory 
$^{4}$ The Chinese University of Hong Kong\\
{\tt\small \{xuyt1,congpsh,yaoych,mayuexin\}@shanghaitech.edu.cn}}

\makeatletter
\let\@oldmaketitle\@maketitle% Store \@maketitle
\renewcommand{\@maketitle}{
   \@oldmaketitle% Update \@maketitle to insert...
 \begin{center}
     \includegraphics[width=2.1\columnwidth]{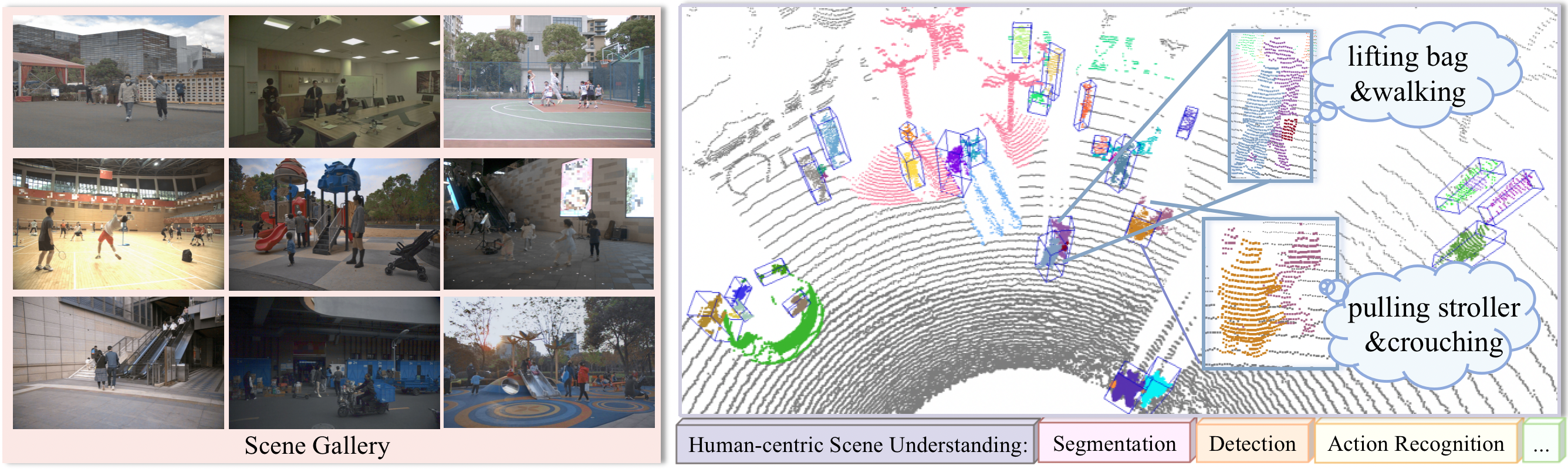}
 \end{center}
   %\vspace{-1ex}
  \refstepcounter{figure}\normalfont Figure~\thefigure. 
  The left shows several scenes captured in HuCenLife, which covers diverse human-centric daily-life scenarios. The right demonstrates rich annotations of HuCenLife, which can benefit many tasks for 3D scene understanding .
  \label{fig:teaser}
  \newline
  }
\makeatother

\maketitle
\ificcvfinal\thispagestyle{empty}\fi

%%%%%%%%% ABSTRACT
\begin{abstract}

% The research of scene understanding based on image and video has developed for a long time. However, the existing data sets mainly focus on narrow and single indoor scenes. For government security or automatic driving scenarios, the existing work cannot accurately locate and analyze multiple people in a large scene at the same time, Image-based methods often fail to work at night, and due to the lack of depth information in the image, it is difficult to distinguish the people who block each other in a scene. However, LiDAR can capture accurate depth information in a long distance and a wide range and LiDAR-based methods is able to accurately locate human position and provide 3D surface point cloud reflecting human posture. We propose HuCenLife, a Human-centric Scene Understanding dataset for 3D Large-scale Scenarios collected by one LiDAR and six cameras, and provides rich annotations for multiple tasks. Based on HuCenLife, we provide three baselines for instance segmentation, 3D human detection and human action recognition. In addition, we also explored the potential application of HuCenLife in other tasks.
Human-centric scene understanding is significant for real-world applications, but it is extremely challenging due to the existence of diverse human poses and actions, complex human-environment interactions, severe occlusions in crowds, etc. In this paper, we present a large-scale multi-modal dataset for human-centric scene understanding, dubbed HuCenLife, which is collected in diverse daily-life scenarios with rich and fine-grained annotations. Our HuCenLife can benefit many 3D perception tasks, such as segmentation, detection, action recognition, etc., and we also provide benchmarks for these tasks to facilitate related research. In addition, we design novel modules for LiDAR-based segmentation and action recognition, which are more applicable for large-scale human-centric scenarios and achieve state-of-the-art performance. The dataset and code can be found at
\url{https://github.com/4DVLab/HuCenLife.git}.

\end{abstract}

%%%%%%%%% BODY TEXT
\section{Introduction}
\label{sec:intro}

% \textcolor{red}{1: why LiDAR is indispensable?  Can not get its necessary from the draft; One possible option is that HuCenLife is the first 3D human centric dataset and our key contribution is to provide the infra for community and facilitate the related research}

% \textcolor{red}{2: Is there any unique task which is relied on LiDAR (or depth information)? We can highlight some specific tasks which is not available in the common datasets}

% \textcolor{red}{3: Compared to STCrowd, highlight the difference of HuCenLife; A teaser is required to show the key contribution or difference; The structure of Introduction can follow the style of STCrowd}

% why human-centric? --> current datasets neglecting? --> how we make the remedy? --> Specific tasks and some novel methods

Human-centric scene understanding in 3D large-scale scenarios is attracting increasing attention~\cite{Dai2022HSC4DH4,cong2022stcrowd,lip,lidarcap}, which plays an indispensable role in human-centric applications, including assistive robotics, autonomous driving, surveillance, human-robot cooperation, \etc. It is often confronted with substantial difficulties since these human-centric scenarios usually have the attributes of various subjects with different poses, fine-grained human-object interactions, and challenging localization and recognition with occlusions. Moreover, current state-of-the-art perception methods heavily rely on large-scale datasets to achieve good performance. Therefore, to promote the research of human-centric scene understanding, the collection of large-scale datasets with rich and fine-grained annotations is required urgently, which is difficult but of great significance.     

% Human-centric understanding in 3D large-scale scenarios is attracting more and more attention~\cite{Dai2022HSC4DH4,STcrowd,lip,lidarcap}.
% Fine-grained segmenting and recognizing human-object interactions is a critical task, which can benefit a variety of real-world applications including robotics, autonomous driving, surveillance, human-machine cooperation, etc.
% However, recognizing and detecting interactions involving both humans and objects in a scene is a challenging task in complex and cluttered environments. Compared with traffic scenarios, human with diverse poses dominate the scenarios and the object distribution tends to follow a long-tailed distribution. Moreover, both human and some human carry-on objects are relatively small and closely located, leading to sparse points in distance and overlapping or stitching points between human and object instances. Furthermore, instances may occlude each other in human-object interaction scenes. These above situation makes the accurate object segmentation and human action recognition quite challenging. 

In previous work, many studies target on the scene understanding based on the input of image or video~\cite{arnab2021vivit,piergiovanni2022rethinking,dosovitskiy2020image,yan2022multiview}, which are not applicable to real-world applications due to the limited 2D visual representations. Afterward, some works pay attention to the static indoor-scene understanding~\cite{dai2017scannet,armeni20163d,bhatnagar2022behave} based on the pre-scanned RGB-D data, which are not suitable for the research of real-time perception. Recently, more and more outdoor multi-modal datasets~\cite{caesar2020nuscenes,waymoopen} are released equipped with LiDAR point clouds. They provide detailed annotations under complex outdoor scenes, while they often focus on the vehicle-dominated traffic environment and neglect the more challenging human-centric daily-life scenarios. Although the dataset STCrowd~\cite{cong2022stcrowd} appears lately, it focuses on the detection task of dense pedestrian scenes, lacking varied human activities and diversified annotations. Consequently, the dataset with rich and fine-grained annotations for human-centric understanding in long-range 3D space is crucial and insufficient.

In this paper, to facilitate the research of human-centric 3D scene understanding, we collect a large-scale multi-modal dataset, namely HuCenLife, by using calibrated and synchronized camera and LiDAR.
Specifically, the dataset captures $32$ multi-person involved daily-life scenes with rich human activities and human-object interactions. Various indoor and outdoor scenarios are both included. For the annotation, we provide fine-grained labels including instance segmentation, 3D bounding box, action categories, and continuous instance IDs, which can benefit various 3D perception tasks, such as point cloud segmentation, detection, action recognition, Human-Object Interaction (HOI) detection, tracking, motion prediction, \etc. In this paper, we provide benchmarks for the former three tasks by executing current state-of-the-art methods on HuCenLife and give discussions for other downstream tasks.

In particular, considering the specific characteristics of human-centric scenarios, we propose effective modules to improve the performance for point cloud-based segmentation and action recognition in the complex human-centric environments. First, we model human-human interactions and human-object interactions and leverage their mutual relationships to benefit the classification of points and instances. Second, to solve the problem of the big scale span of objects in daily-life scenarios, we exploit multi-resolution feature extraction strategy to aggregate global features and local features hierarchically so that small objects can be better attended.
%In this paper, we mainly focus on three human-centric perception tasks, including human-centric point cloud segmentation, human-centric 3D detection and human-centric action recognition. Moreover, we further propose the Human-Human-Object-Interaction (HHOI) module and  Ego-Neighbour Feature
%Interaction (ENFI) to model the complex interactions in human-centric scenarios, and design the Hierarchical Point Feature Extraction (HPFE) capture both local features and global features in hierarchical resolutions. 
We evaluate our methods and conduct extensive experiments on HuCenLife. Several ablation studies are also conducted to demonstrate the effectiveness of each module and good generalization capability. Our contributions are summarized as follows:

% Human behavior in human-centric scenarios is often complex and closely related to surrounding objects and environments. We propose the Human-Human-Object-Interaction(HHOI) module to obtain a better instance segmentation result. Our module integrates the presentation between diverse human and learn a weighted feature guided by human-centric feature to enhance the object feature. 
% Human action recognition also depends on interacted object and the behaviours of surrounding people. Considering the different size of interacted objects, we design Hierarchical Point Feature Extraction(HPFE), to capture both local features and global features at various resolutions. To extract the human-human interaction, Ego-Neighbour Feature Interaction(ENFI) is proposed to combine the ego-human feature with the neighbours through a cross-attention mechanism. 
% We evaluate our method on HuCenLife and achieve state-of-the-art performance. Several ablation studies also conducted to demonstrate our effectiveness and good generalization capability. Our contributions are summarized as follows:
\begin{enumerate}
    \item We introduce HuCenLife, the first large-scale multi-modal dataset for human-centric 3D scene understanding with rich human-environment interactions and fine-grained annotations. 
    \item HuCenLife can benefit various human-centric 3D perception tasks, including segmentation, detection, action recognition, HOI, tracking, motion prediction, etc. We provide baselines for three main tasks to facilitate future research.
    \item Several novel modules are designed by incorporating fine-grained interactions and capturing features at various resolutions to promote more accurate perception in human-centric scenes.
    
\end{enumerate}

\section{Related Work}
\label{sec:related}
\subsection{Datasets for 3D Scene Understanding}
The RGB-D datasets of indoor scenes dominate the early scene understanding task. ScanNet~\cite{dai2017scannet,armeni20163d} focuses on object surface reconstruction and semantic segmentation, providing dense and rich annotations for various indoor objects. NTU RGB+D~\cite{shahroudy2016ntu} is a human action recognition dataset with corresponding skeleton and action labels. Behave~\cite{bhatnagar2022behave} concentrates on human-object interaction with human SMPL models and interactive objects annotations. It can be found that outdoor scenarios are not well explored.
Recently, the community has paid attention to traffic scenes for autonomous driving and collect several outdoor multi-modal datasets. KITTI~\cite{geiger2012we}, nuScenes~\cite{caesar2020nuscenes} and Waymo~\cite{sun2020scalability} provide 3D bounding boxes for traffic participants and ~\cite{caesar2020nuscenes,behley2019semantickitti} also offer point-wised semantic segmentation labels. However, these datasets are all vehicle-dominated and neglect human-centric scenarios.
STCrowd~\cite{cong2022stcrowd} mainly concentrates on the crowds on campus but lacks the fine-grained segmentation labels and complex human-environment interactions.
In order to facilitate the research of human-centric 3D scene understanding, we collect HuCenLife, a multi-modal dataset with various scenarios in human daily life.
\subsection{Point Cloud-based Segmentation}
% outdoor 
Most outdoor point cloud segmentation methods mainly focus on point cloud representations. Point-based methods ~\cite{qi2017pointnet,qi2017pointnet++,yue2019dynamic,thomas2019kpconv} make the operation on unordered point cloud directly. Voxel-based methods ~\cite{choy20194d,graham20183d} utilize efficient sparse convolution to reduce the time complexity. PolarNet~\cite{zhang2020polarnet} and Cylinder3D~\cite{zhu2021cylindrical} further consider the non-uniform LiDAR point clouds characteristics and point distribution, and divide the points under the polar coordinate system. ~\cite{hong2021lidar} adopts the cylinder convolution and proposes a dynamic shifting network for instance prediction.
These methods are mainly focusing on automatic driving scenes, while neglecting the counterpart in human-centric scenarios with complex human-object interactions and challenging occlusions.

Another line of segmentation, namely point cloud instance segmentation, also embraces great progress, which can be mainly divided into proposal-based methods and grouping-based methods. Previous proposal-based methods ~\cite{yi2019gspn,engelmann20203d,yang2019learning} regard the instance segmentation as a top-down  pipeline, which first generate proposals and then segment the objects within the proposals. Grouping-based methods ~\cite{jiang2020pointgroup,vu2022softgroup,han2020occuseg,chen2021hierarchical,he2021dyco3d,wu20223d} adopt the bottom-up strategy. PointGroup ~\cite{jiang2020pointgroup} aggregates points from original and offset-shifted point sets. DyCo3D ~\cite{chen2021hierarchical} and DKNet~\cite{wu20223d} encode instances into kernels and propose dynamic convolution kernels and then merge the candidates. Considering the imprecise bounding box prediction in proposal-based methods for refinement and the time-consuming aggregation in grouping methods, 
~\cite{schult2022mask3d,sun2022superpoint} take each object instance as an instance query and design a query decoder with transformers. However, these methods are applied to structured indoor instances without human involvement and human-environment interactions. Our dataset and proposed method target more on human-human and human-object interactions in large-scale human-centric scenes.

\subsection{LiDAR-based 3D Detection}
% Ranged imaged based:
% point based:
% more popular voxel based {anchor free(center based) anchor based{Ted Voxel RCNN}}

As the mainstream of 3D perception task, 3D detection task has been fully explored, which can be grouped via the point encoding strategies. First, point-based methods ~\cite{IA-SSD,SASA,DeepHoughVoting,pointRCNN,3DSSD} extract the geometry information from raw points with sampling and grouping. ~\cite{tianfully,SqueezeSeg,bewley2021range,sun2021rsn,lang2019pointpillars,fan2021rangedet} transform point cloud into range images for detection. Second, voxel-based methods ~\cite{voxelnet,TED,MGTANet,deng2022vista,voxel-rcnn,centerpoint,yan2018second} convert raw point clouds to regular volumetric or pillar representations and adopt voxel-based feature encoding. 
Third, hyper-fusion methods ~\cite{peng2022pv,jiang2021vic,shi2020pv,yang2019std,chen2019fast,centerformer} take advantage of both voxels and points and fuse them together to model the hyper encoding. In this paper, we test them on the proposed HuCenLife dataset to provide the benchmark and offer the comprehensive analyses and comparison.  

% 3D detection aims to predict the 3D bounding box and the category of each object in the given point cloud. 3D detection can be roughly classified by their point encoding methods. For point-based method ~\cite{IA-SSD,SASA,DeepHoughVoting,pointRCNN,3DSSD}, they extract information directly from points' geometry. Different from point-based method, ~\cite{tianfully,SqueezeSeg,bewley2021range,sun2021rsn,lang2019pointpillars,fan2021rangedet} transform point cloud into range image and use it as input. Based on the idea of ~\cite{voxelnet}, voxel-based method ~\cite{centerformer,TED,MGTANet,STcrowd,deng2022vista,voxel-rcnn,centerpoint,yan2018second} emerges. Space is divided into voxels, thus convolution can be applied. There're also point-voxel based method ~\cite{peng2022pv,jiang2021vic,shi2020pv,yang2019std,chen2019fast} who try to take advantage of both voxel based and point based. While voxel-based best remain to be the most popular method. Among voxel-based, some of them~\cite{yan2018second,voxel-rcnn,TED}  are anchor-based. For the others, most of them~\cite{centerformer,MGTANet,STcrowd,deng2022vista,centerpoint} are based on heatmap to estimate the label and location of the target. Most of these algorithm are designed for auto-mobile and test on the datasets for auto-mobile who focus primarily on verticals. Some of them ~\cite{STcrowd} have paid close attention to pedestrian. However even them have neglect the variety action human can have. We are focus on the human with variety of pose and actions in large-scale scenes.
\subsection{Action Recognition}
% The research of human action recognition is mainly concentrated in the field of video classification.  
Recently, transformer-based methods have dominated the field of action recognition~\cite{li2022uniformerv2,fang2022eva}. Many variants based on ViT~\cite{dosovitskiy2020image} have been proposed to explore the potential of transformer in video classification, where
ViViT~\cite{arnab2021vivit} extends the two-dimensional patch to three-dimensional tube to model the temporal relation, MTV~\cite{yan2022multiview} divides the tube with different time scales to extract the action features with different amplitude of change over time, and TubeViT~\cite{piergiovanni2022rethinking} further samples various sized 3D space-time tubes from the video to generate learnable tokens.
% TubeViT  sample various sized 3D space-time tubes from the video to generate learnable tokens, which  can largely reduce computation burden for local relation and flexibly build long-range token dependencies from distant frames simultaneously. 
However, the common action recognition \cite{song2015sun,shahroudy2016ntu} is annotated in image-level and lacks of instance-level labels, causing these methods hard to be applicable in complex 3D scenarios. In this paper, we introduce point cloud-based instance action recognition task in large-scale scenes and collect the HuCenLife dataset equipped with various instances with different poses and motions, to make the basis for research community.

\section{HuCenLife Dataset}
\label{sec:dataset}

\begin{table*}[ht]
\centering
\caption{Comparison with related datasets for 3D scene understanding. There are some abbreviations, where ``pc'' denotes LiDAR point cloud, ``ins. seg.'' means instance segmentation, ``bbx'' is bounding box, and ``inter. obj.'' denotes objects having interactions with humans.}
\label{tab:compare}
\resizebox{\linewidth}{!}{
\begin{tabular}{c|c|c|c|c|c|cc|ccc|cc}
\hline
      \rowcolor{blue!5}   Dataset     & Data   &  LiDAR & Point Cloud
&Person &Person &\multicolumn{2}{c|}{Scenes} &\multicolumn{3}{c|}{Annotation Content}& \multicolumn{2}{c}{Annotation Targets} \\ \cline{7-13}   

\rowcolor{blue!5} &Modality & Beam&Frame&Number&Per Frame& indoor & outdoor & ins. seg.& 3D bbx& action& multi-person& inter. obj.
\\\hline
ScanNet\cite{dai2017scannet}               &RGBD         & -  &-&-&-& \textcolor{green}{\Checkmark}& \textcolor{red}{\XSolidBrush}    &  \textcolor{green}{\Checkmark}       &    \textcolor{green}{\Checkmark}      & \textcolor{red}{\XSolidBrush}              & \textcolor{red}{\XSolidBrush}  & \textcolor{red}{\XSolidBrush}  \\\hline
S3DIS\cite{armeni20163d}&RGBD  
&-&-&-& -  & \textcolor{green}{\Checkmark}& \textcolor{red}{\XSolidBrush}    &  \textcolor{green}{\Checkmark}       &    \textcolor{green}{\Checkmark}      & \textcolor{red}{\XSolidBrush}              & \textcolor{red}{\XSolidBrush}  & \textcolor{red}{\XSolidBrush}  \\\hline
SUN RGB-D\cite{song2015sun}&RGBD&-
&-&-&-&\textcolor{green}{\Checkmark} &\textcolor{red}{\XSolidBrush}    & \textcolor{red}{\XSolidBrush}    &\textcolor{green}{\Checkmark}   & \textcolor{red}{\XSolidBrush}         &\textcolor{red}{\XSolidBrush}    & \textcolor{red}{\XSolidBrush}
\\\hline
NTU RGB+D\cite{shahroudy2016ntu}                &RGBD        &-  &-&-&-
&\textcolor{green}{\Checkmark} &\textcolor{red}{\XSolidBrush}    & \textcolor{red}{\XSolidBrush}       & \textcolor{red}{\XSolidBrush}       &\textcolor{green}{\Checkmark}  &\textcolor{red}{\XSolidBrush}    & \textcolor{red}{\XSolidBrush}     \\\hline

BEHAVE\cite{bhatnagar2022behave}          &RGBD       &-   &
15.8k& 15.8k&1
&\textcolor{green}{\Checkmark} &\textcolor{red}{\XSolidBrush}    &   \textcolor{green}{\Checkmark}     &  \textcolor{red}{\XSolidBrush}      & \textcolor{green}{\Checkmark}      &\textcolor{red}{\XSolidBrush}    &\textcolor{green}{\Checkmark}          \\\hline

SemanticKITTI\cite{behley2019semantickitti}      &pc      & 64 & 43k&9.7k&0.2&\textcolor{red}{\XSolidBrush}  & \textcolor{green}{\Checkmark} &   \textcolor{green}{\Checkmark}     &  \textcolor{red}{\XSolidBrush}         &\textcolor{red}{\XSolidBrush}  &    \textcolor{green}{\Checkmark}           & \textcolor{red}{\XSolidBrush}       \\\hline

KITTI\cite{geiger2012we}      &image\&pc      & 64 & 15k &4.5k&0.3&
\textcolor{red}{\XSolidBrush}  & \textcolor{green}{\Checkmark}  &  \textcolor{green}{\Checkmark} &   \textcolor{green}{\Checkmark}            &\textcolor{red}{\XSolidBrush}  &    \textcolor{green}{\Checkmark}           & \textcolor{red}{\XSolidBrush}       \\\hline

Waymo\cite{sun2020scalability}         &image\&pc   & 64 &  230k& 2.8M&12&
\textcolor{red}{\XSolidBrush}  & \textcolor{green}{\Checkmark}&   \textcolor{green}{\Checkmark}     &   \textcolor{green}{\Checkmark}     &\textcolor{red}{\XSolidBrush}  &    \textcolor{green}{\Checkmark}           & \textcolor{red}{\XSolidBrush}       \\\hline

nuScenes\cite{caesar2020nuscenes}         &image\&pc     & 32 & 40k &208k&5&\textcolor{red}{\XSolidBrush}  & \textcolor{green}{\Checkmark}&   \textcolor{green}{\Checkmark}     &   \textcolor{green}{\Checkmark}     &\textcolor{red}{\XSolidBrush}  &    \textcolor{green}{\Checkmark}           & \textcolor{red}{\XSolidBrush}       \\\hline

STCrowd\cite{cong2022stcrowd}           &image\&pc    & 128  & 11k &219k&20&\textcolor{red}{\XSolidBrush}  & \textcolor{green}{\Checkmark} & \textcolor{red}{\XSolidBrush}        & \textcolor{green}{\Checkmark}        & \textcolor{red}{\XSolidBrush}  &  \textcolor{green}{\Checkmark}        & \textcolor{red}{\XSolidBrush}         \\\hline

\textbf{HuCenLife}          &image\&pc      &128 & 6.1k& 65k&11&\textcolor{green}{\Checkmark}   & \textcolor{green}{\Checkmark}  & \textcolor{green}{\Checkmark}       & \textcolor{green}{\Checkmark}  & \textcolor{green}{\Checkmark} &  \textcolor{green}{\Checkmark} & \textcolor{green}{\Checkmark}  \\ \hline
\end{tabular}
}
\vspace{-2ex}
\end{table*}

HuCenLife is the first dataset that emphasizes human-centric 3D scene understanding, containing indoor and outdoor daily-life scenes with rich annotations of human activities, human-human interactions, and human-object interactions, which facilitates the development of intelligent security, assistive robots, human-machine cooperation, \etc.
%HuCenLife focuses on human-centric 3D scene understanding and captures diverse daily-life scenes, which contains rich human-human interactions and human-object interactions in large-scale indoor or outdoor scenarios. 
In this section, we first introduce the data acquisition in Sec.\ref{subsec:Acquisition}, and then provide important annotation statistics in Sec.\ref{subsec:Statistics}, and finally highlight the novelties of HuCenLife by comparing with existing influential datasets in Sec.\ref{subsec:Characteristic}.

\subsection{Data Acquisition}
\label{subsec:Acquisition}
\begin{figure}[t]
    \centering
    \includegraphics[width=1\columnwidth]{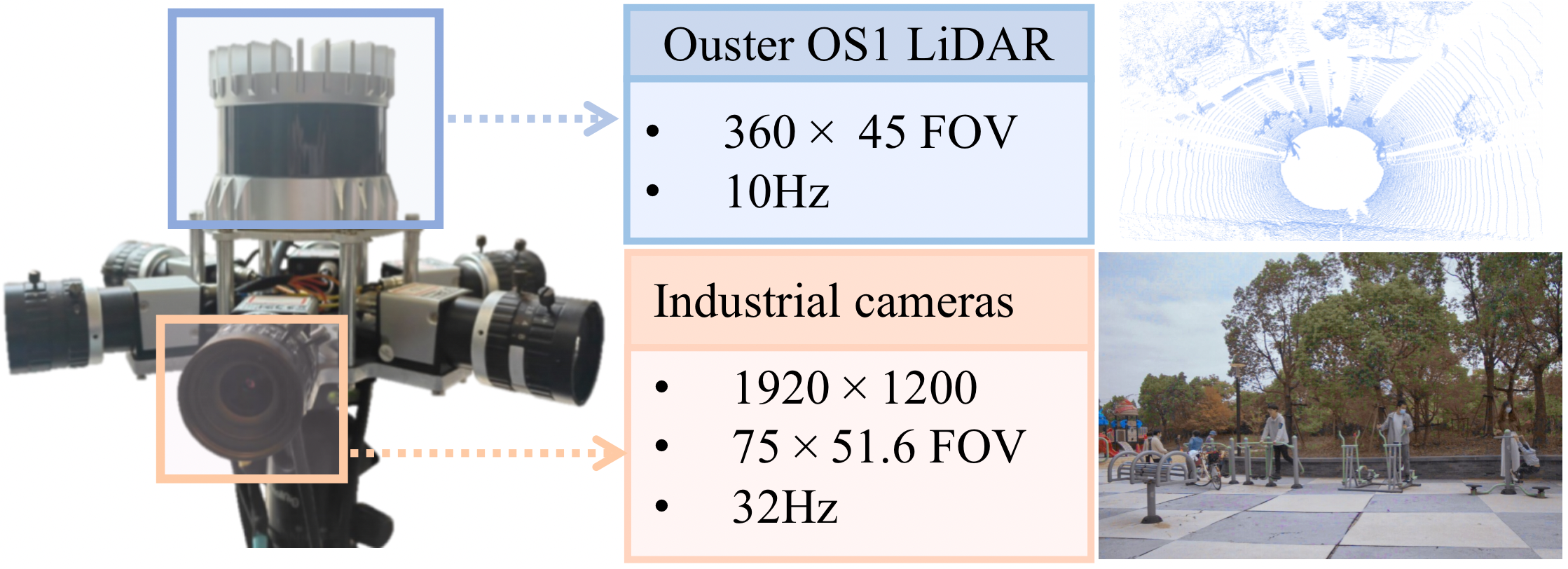}
    \caption{Sensor setup for data collection. }
    \label{fig:sensor}
    \vspace{-2ex}
\end{figure}

To collect the dataset, we built a Visual-LiDAR Capture System, which mainly consists of one 128-beam Ouster-OS1 LiDAR and six industrial cameras in a circle, as Fig ~\ref{fig:sensor} shows. All sensors are tied in fixed positions on the bracket with mechanical synchronization. The LiDAR has a $360^{\circ}$ horizon field of view (FOV) $\times 45^{\circ}$ vertical FOV, and each camera has a $75^{\circ} \times 51.6^{\circ}$ FOV with $1920\times1200$ image resolution. For our equipment, LiDAR captures raw point cloud in $10$Hz and camera takes pictures in $32$Hz.

% \begin{figure}[t]
%     \centering
%     \includegraphics[width=1\columnwidth]{pic/dataset_scenes.png}
%     \caption{Diverse scenes in HuCenLife, containing indoor and outdoor large-scale scenarios in day and night with rich human activities, human-human interactions, and human-object interactions. }
%     \label{fig:dataset_scenes}
% \end{figure}

\subsection{Annotation}
\label{subsec:Statistics}

We manually annotated all humans and these objects with interactions with humans in LiDAR point cloud by referring to the synchronized image. We select one frame per second for labeling and finally obtain $6,185$ frames ($103$ minutes) of annotated LiDAR point cloud. For each target, we provide four kinds of annotations, \ie, point cloud-based instance segmentation, 3D bounding box, human action classification, and tracking ID across consecutive frames, like Fig ~\ref{fig:teaser} shows. In HuCenLife, there are $65,265$ human instances in total, including $58,354$ adults and $6,911$ children, and $31,303$ human-interacted objects. There are $20$ categories of objects and $12$ kinds of human actions. Specifically, the HuCenLife dataset is collected in $15$ distinguished locations with $32$ human-centric daily-life scenes, including playground, shopping mall, campus, park, gym, meeting room, express station, \etc. For each scene, there are $11$ persons on average with multiple interacted objects, and for some complex scenes, there are about $70$ persons. The diverse density distributions in HuCenLife bring challenges for related research.
%Because the visual-LiDAR device is usually settled beside the scene to eliminate its affect on peoples' normal activities, we only annotate $180^{\circ}$ view of LiDAR point cloud and corresponding images of three cameras (where different scenes vary in cameras), which covers pertinent contents of scenes. 
More detailed annotation introductions are in the supplementary material.

\subsection{Characteristics}
\label{subsec:Characteristic}
We introduce the basic information of HuCenLife and compare it with related popular datasets in Table ~\ref{tab:compare}. In particular, we conclude four highlights of our dataset below.
%from five main aspects, including the modality of acquisition data, sensor configuration, data-collection scenes, annotation content, and annotation targets. In particular, we  

\textbf{Large-scale Dynamic Scenarios.} 
Benefiting from the long-range-sensing and light-independent properties of LiDAR, HuCenLife contains data of diverse large-scale scenes day and night. Unlike indoor datasets~\cite{dai2017scannet} where the scene is pre-scanned and has only static objects, HuCenLife provides online captured multi-modal visual data of dynamically changing scenes with dynamic people, objects, and background. Furthermore, the density of humans and objects is changing from a few to dozens in distinct scenes. The visual data in such diverse dynamic scenarios has huge significance for developing mobile robots.  

\textbf{Abundant Human Poses.} 
Different from current traffic or crowd datasets~\cite{sun2020scalability,caesar2020nuscenes,cong2022stcrowd}, where people only act as pedestrians walking or standing on the road, HuCenLife pays attention to daily-life scenarios, where people have rich actions, such as doing exercise, crouching down, dancing, running, riding, \etc. In particular, HuCenLife contains thousands of children samples, which are never concerned in previous datasets. Such complex scenarios with high-degree freedom of human poses bring challenges for accurate perception and recognition.  

\textbf{Diverse Human-centric Interactions.} 
Apart from abundant self-actions of humans, HuCenLife also includes rich human-human interactions (hugging, holding hands, holding a baby, \etc.) and human-object interactions (riding a bike, opening the door, carrying a box, \etc.). What's more, there are some extremely complex human-human-object interactions, such as playing basketball, having a meeting in a room, \etc., which require the participation of multiple persons and objects. HuCenLife is unique for containing diversified interaction data in a variety of scenes, which is significant for the research of human-machine cooperation and boosts the development of service robots.

\textbf{Rich Annotations.} 
HuCenLife provides rich fine-grained annotations, which can benefit many perception tasks, such as point cloud segmentation, 3D detection, 3D tracking, action recognition, HOI, motion prediction, \etc. In particular, due to complex scene contents, the annotation process of HuCenLife is much more difficult than others. A well-trained annotator usually spends $25$min on average for labeling one frame of LiDAR point cloud in our dataset.

\subsection{Privacy Preservation}
We strictly obey the privacy-preserving rules. We mask all sensitive information, such as the faces of humans and locations, in RGB images. LiDAR point clouds without any texture and facial information naturally protect the privacy.

%Previous datasets for 3D scene understanding focus on two main scenarios, including the indoor static scenes with many types of furniture and outdoor traffic scenes with various vehicles and road structures. HuCenLife is the first dataset to pay attention to large-scale human-centric daily life scenarios, including indoor and outdoor scenes in day and night with rich human activities, human-human interactions, and human-object interactions, which are significant for the development of intelligent security, assistive robots, human-machine cooperation, etc. We compare with related popular datsets in Tab.\ref{tab:compare} from five main aspects, including the modality of acquisition data, sensor configuration, data-collection scenes, annotation content, and annotation targets. We can see that HuCenLife provides totally different scene data and more fine-grained annotations, which can benefit many 3D perception tasks, such as point cloud segmentation, detection, action recognition, HOI, motion prediction, etc. In particular, due to complex scene contents, the annotation process of HuCenLife is much more difficult than others.

\section{Various Downstream Tasks}

As mentioned above, our dataset can benefit numerous human-centric 3D perception tasks. We conduct three main tasks on HuCenLife based on the LiDAR point cloud, including human-centric instance segmentation, human-centric 3D detection, and human-centric action recognition, and provide the baseline methods. Particularly, novel methods are proposed for instance segmentation and action recognition, respectively, to tackle the difficulties of large-scale human-centric scenarios. In what follows, we present details of these tasks with extensive experiments in order.
%-------------------------------------------------------------------------
\section{Human-centric Instance Segmentation}
\label{sec:insseg}

\begin{table*}[ht!]\small
\caption{Instance segmentation results on HuCenLife dataset.} \label{tab:test}
	\setlength{\tabcolsep}{1.8mm}
 \resizebox{\linewidth}{!}{      
\begin{tabular}{c|c|c|c|c|c|c|c|c|c|c|c|c|c|c|c|c|c|c|c|c|c|c|c}
\hline
                  & \rotatebox{90}{person} & \rotatebox{90}{motorbike} & \rotatebox{90}{table} & \rotatebox{90}{box}   & \rotatebox{90}{cart}  & \rotatebox{90}{seesaw} & \rotatebox{90}{basketball} & \rotatebox{90}{fitness equ} & \rotatebox{90}{cabinet} & \rotatebox{90}{baby}  & \rotatebox{90}{blackboard} & \rotatebox{90}{staircase} & \rotatebox{90}{slide} & \rotatebox{90}{scooter} & \rotatebox{90}{computer} & \rotatebox{90}{backpack} & \rotatebox{90}{obj in hand} & \rotatebox{90}{chair} & \rotatebox{90}{spring car} & \rotatebox{90}{ground} & mIOU  & AP50 & AP25 \\\hline
% Softgroup          & 78.3  & 22.2      & 3.1  & 42.0 & 24. & 65.7  & 13.4      & 24.6             & 1.7    & 37.6 & 90.6      & 11.5     & 96.0            & 25.6         & 35.5    & 30.3    & 6.8        & 6.9  & 34.4      & 97.1  & 37.4 &      &      \\\hline

Voxel-DSNet \cite{hong2021lidar}   &66.9&16.1&20.7&   22.6  &16.3 &12.6      &5.4&11.9&1.1&25.7&58.3&8.5&72.7&33.3&24.6&20.7&1.6&8.6&3.1&97.9 &    26.4 &2.6&7.1\\\hline
Cylinder-DSNet \cite{hong2021lidar}&72.3&12.9&{23.8}&28.6&18.9&25.2&5.8&4.7&6.8&23.4&90.2&{21.4}&67.9&37.2&15.5&23.9&3.5&{14.1}&2.8&97.9&29.8&1.2& 7.6\\\hline
DKNet  \cite{wu20223d}
&75.6&{52.7}&5.3&26.3&35.8&65.6&0.0&14.6&0.6&39.7&93.9&0.0&95.1&{48.5}&13.1&9.8&{14.6}&8.1&3.4&{98.0}& 35.0
&11.1 & 14.0 \\\hline
SoftGroup \cite{vu2022softgroup}& 80.0  & 32.6      & 4.4  & 38.2 & 20.6 & 60.7  & 8.3       & 25.2             & 3.2    & {42.5} & {95.5}      & 1.0      & 95.8            & 24.6         & 27.5    & {34.0}    & 7.1        & 7.6  & 29.0      & 96.2  & 36.7 & 32.5 & 38.2 \\\hline
% Ours(w/o opt)     & 81.6  & 30.7      & 8.6  & 40.0 & 41.1 & 67.2  & 20.7      & 26.4             & 3.8    & 27.7 & 88.1      & 2.5      & 96.4            & 38.3         & 35.6    & 33.1    & 7.4        & 9.5  & 24.7      & 96.7  & 39.0 & 33.0 & 38.8 \\\hline

\rowcolor{gray!10} Ours &{82.7}&46.4&6.4&{39.7}&{51.1}&{69.4}&{15.3}&{29.6}&3.0&40.0&89.4&1.2&{96.8}&35.6&{29.2}&28.4&6.8&10.6&{32.3}&96.9& \textbf{40.5}&\textbf{35.6}&\textbf{40.4}\\\hline\hline
Ours + PointPainting
&79.8&30.7&16.2&42.5&{47.6}&53.4&8.1&21.7&{3.9}&32.8&82.3&0.0&95.6&34.2&19.6&25.3&{11.9}&{19.7}&30.0&96.4&37.6&28.9&34.8\\\hline 
w/o HHIO     & 79.5 &15.2 &{17.4} &32.9 &31.6 &56.6 &7.1 &26.1 &1.8 &35.0 &92.8 &0.6 &95.8 &22.0 &26.7 &30.2 &9.5 &19.0 &29.0 &{97.1} & 36.3 & 25.0 &  31.6  \\\hline
Ours + LocalFusion      &{81.8}&{46.8}&2.0&{46.2}&36.8&{74.7}&13.2&{28.5}&0.4&{37.3}&{93.8}&{2.3}&{96.5}&{35.2}&{37.0}&27.8&8.6&9.9&26.0&96.5 &40.1    &  36.9    & 42.0  \\\hline 
w/o HHIO      & 80.7  & 39.3      & 2.3  & 41.9 & 26.6 & 73.1  & {13.9}      & 23.2             & 2.2    & 35.4 & 92.2      & 0.0      & 94.4            & 24.5         & 29.7    &{ 31.4}    & 7.3        & 13.6 & {36.1}      & 95.9  & 38.2 &  36.6    &  41.6  \\\hline 
\end{tabular}
}
%\vspace{-2ex}
\end{table*}

\begin{table*}[ht!]\tiny
\caption{Semantic segmentation results on BEHAVE dataset.} 
\centering
\label{tab:behave}
	\setlength{\tabcolsep}{1.8mm}
  \resizebox{\linewidth}{!}{  
\begin{tabular}{c|c|c|c|c|c|c|c|c|c|c|c|c|c|c|c|c|c|c|c|c|c|c}
\hline
                  & \rotatebox{90}{person} & \rotatebox{90}{backpack} & \rotatebox{90}{basketball} & \rotatebox{90}{boxlarge}   & \rotatebox{90}{boxlong}  & \rotatebox{90}{boxmedium} & \rotatebox{90}{boxsmall} & \rotatebox{90}{boxtiny} & \rotatebox{90}{chairblack} & \rotatebox{90}{chairwood}  & \rotatebox{90}{keyboard} & \rotatebox{90}{monitor} & \rotatebox{90}{container} & \rotatebox{90}{stool } & \rotatebox{90}{suitcase} & \rotatebox{90}{tablesmall} & \rotatebox{90}{tablesquare} & \rotatebox{90}{toolbox} & \rotatebox{90}{trashbin} & \rotatebox{90}{yogaball} &\rotatebox{90}{yogamat} & mIOU   \\\hline
SoftGroup\cite{vu2022softgroup} &96.8&71.8&54.9&83.7&54.4&56.3&40.4&21.9&85.6&82.4&27.1&67.3&71.7&83.5&77.7&82.5&90.5&44.5&64.7&89.0&70.5 & 67.5\\\hline
Ours &97.0&72.4&61.3&86.6&62.2&57.3&45.6&33.4&87.7&83.3&30.8&72.1&73.8&84.2&76.0&86.6&91.9&49.2&66.9&89.0&76.4 & \textbf{70.7}\\\hline
\end{tabular}
}
\vspace{-3ex}
\end{table*}

For LiDAR point cloud-based semantic instance segmentation, the input is expressed as $ P \in \mathcal{R}^{N \times 4}$, which involves $N$ points with the 3D location and reflection intensity $(x, y, z, r)$. The task is to assign each point to a category and then output a set of object instances with their corresponding semantic labels.

\subsection{Method}

For human-centric scenes, people have diverse pose types and may stay together with occlusions. Moreover, some objects are relatively small and closely located to the person, causing overlapping or stitching points with humans and bringing difficulties in distinguishing from the person.
% For human-object integration scenes, person with diverse pose types dominates the dataset and the objects are close to the person. While the categories of the object have a long-tailed distribution and some small things are hard to distinguish and separated from person. Moreover, the person may walking together with occlusion, resulting in partial points and the person interaction is also worth analyzing.  
%Existing methods for outdoor traffic scenes and indoor scenes neglect the fine-grained segmentation such as small object closed with person and the interaction between humans and human-objects. 
% mainly focusing on the feature representation and efficient partition, while neglecting the fine-grained segmentation such as small object closed with person. Several indoor instance segmentation methods pay more attention on grouping and clustering the instance, while they are applied on structured furniture with dense RGB-D representations without consider the participation of human and the sparsity of LiDAR. 
To tackle these problems, we propose a Human-Human-Object Interaction(HHOI) module, shown in Figure \ref{fig:segnetwork}. The model first extracts the human-human interaction feature with attention strategy so that humans can be more accurately recognized even with partial point cloud in occluded scenes.
Then, it uses human-centric features to guide the network automatically to learn a weighted feature to pay attention to interactive objects, which can benefit capturing fine-grained semantic information.

\subsubsection{Human-Human-Object-Interaction Module}
\begin{figure}[t]
    \centering
\includegraphics[width=0.98\columnwidth]{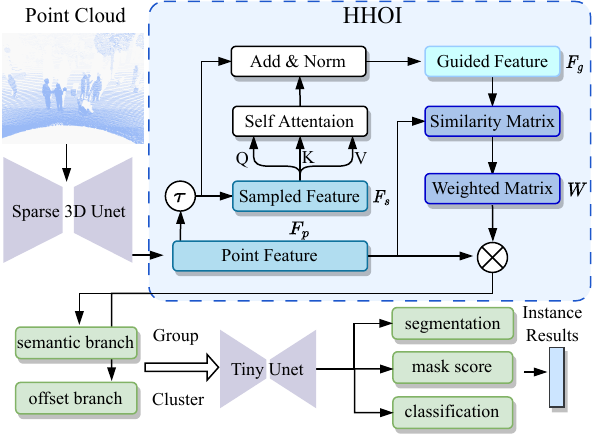}
    \caption{ The architecture of our segmentation method. Especially, the HHOI module extracts the correlation within different persons and the human-object relationships, which can benefit the point-wise and instance-wise classification.
    %The point cloud is fed in Sparse 3D Unet to extract the point feature. Then, the HHOI module extracts the correlation within different persons and the human-object relationships. After that, two branches are constructed to output the point-wise semantic scores and offset vectors and we obtian the final results through a tiny Unet.
    }
    \label{fig:segnetwork}
    \vspace{-3ex}
\end{figure}

% \subsubsection{Human-Human-Object-Interaction Module}
% The objects are in close contact with person, so we consider the human-guided feature for a better feature representation.
As shown in Figure \ref{fig:segnetwork}, we utilize a sparse 3D Unet to get $D$ dimensional point feature $F_p \in \mathcal{R}^{N \times D}$. Then, human-human interacted features are extracted through a transformer mechanism. 
We get the semantic score $Y = softmax(MLP(F_p)) = \{y_{i,c}\}^{N\times C}$ for each point, where $C$ is the class number. And then we select $M$ points with the confidence of belonging to person class higher than the threshold $\tau$. We further apply the triplet Q, K, V attention layer to extract correlations among different sampled person features $F_s$ and obtain the final human-guided feature:
$$f_{attention} = softmax(\frac{QK^T}{\sqrt{D}})V,$$
$$F_g = LN(f_{attention} + FFN(f_{attention})),$$
where $LN$ is layer normalization and $FFN$ is the feed-forward neural network~\cite{vaswani2017attention}.
Then, we use human-guided feature to extract human-object interaction for fine-grained object segmentation.
The similarity weighted matrix $W = softmax(F_p F_g^T)$ is computed to enhance the features of objects that people interact with. We multiply the weighted matrix with point features to obtain the final weighted features.
In this way, the model adaptively learns human-related representations and enhances the object feature with the guidance of high-confidence human features.

\subsubsection{Point-wise Prediction and Refinement}
Taking the weighted features as input, the semantic branch and offset branch apply two-layer MLP and output the semantic scores $S\in \mathcal{R}^{N \times K}$ and offset vectors  $O\in \mathcal{R}^{N \times K}$  from the point to the instance center, respectively. The weighted cross-entropy loss $\mathcal{L}_{\text {semantic}}$ and $L_1$ regression loss $\mathcal{L}_{\text {offset}} $ are used to train the semantic and offset branches. After that, we follow the refinement stage in SoftGroup~\cite{vu2022softgroup}, where point-level proposals are fed into a tiny-unet to predict classification scores, instance masks, and mask scores to generate the final instance results. Specifically, the classification branch predicts the category scores $c_k$ for each instance. The segmentation branch utilizes a point-wise MLP to predict an instance mask $m_k$ for each instance proposal. Mask scoring branch estimates the IoU between the predicted mask and the ground truth for each instance. We train each branch with cross-entropy loss $\mathcal{L}_{\text {class}}$, binary cross-entropy loss $\mathcal{L}_{\text {mask}}$, and $l_2$ regression loss $\mathcal{L}_{\text{mask score}}$. And the total loss is the sum of all above losses.
%$\mathcal{L} = \mathcal{L}_{\text {semantic}}+\mathcal{L}_{\text {offset}}+\mathcal{L}_{\text {class}}+\mathcal{L}_{\text {mask}}+\mathcal{L}_{\text{mask score}}.$
% ------------ move to supp
% $$
% L_{\text {semantic }}=\frac{1}{N} \sum_{i=1}^N \operatorname{CE}\left(\boldsymbol{s}_i, s_i^*\right),$$
% $$
% L_{\text {offset }}=\frac{1}{\sum_{i=1}^N \mathbb{I}_{\left\{\boldsymbol{p}_i\right\}}} \sum_{i=1}^N \mathbb{I}_{\left\{\boldsymbol{p}_i\right\}}\left\|\boldsymbol{o}_i-\boldsymbol{o}_i^*\right\|_1,
% $$
% $$
% L_{\text {class }}=\frac{1}{K} \sum_{k=1}^K \mathrm{CE}\left(\boldsymbol{c}_k, c_k^*\right), 
% $$
% $$
% L_{\text {mask }}=\frac{1}{\sum_{k=1}^K \mathbb{I}_{\left\{\boldsymbol{m}_k\right\}}} \sum_{k=1}^K \mathbb{I}_{\left\{\boldsymbol{m}_k\right\}} \mathrm{BCE}\left(\boldsymbol{m}_k, \boldsymbol{m}_k^*\right), 
% $$
% $$
% \mathcal{L}_{\text{mask score}}=\frac{1}{\sum_{k=1}^{N_{g t}} \mathbb{I}_{\left\{iou_k\right\}}} \sum_{k=1}^{N_{g t}} \mathbb{I}_{\left\{iou_k\right\}}\left\|iou_k-iou_k^*\right\|_2
% $$
% where $*$ denotes the ground truth.

\begin{figure}[t]
    \centering
    \includegraphics[width=1\columnwidth]{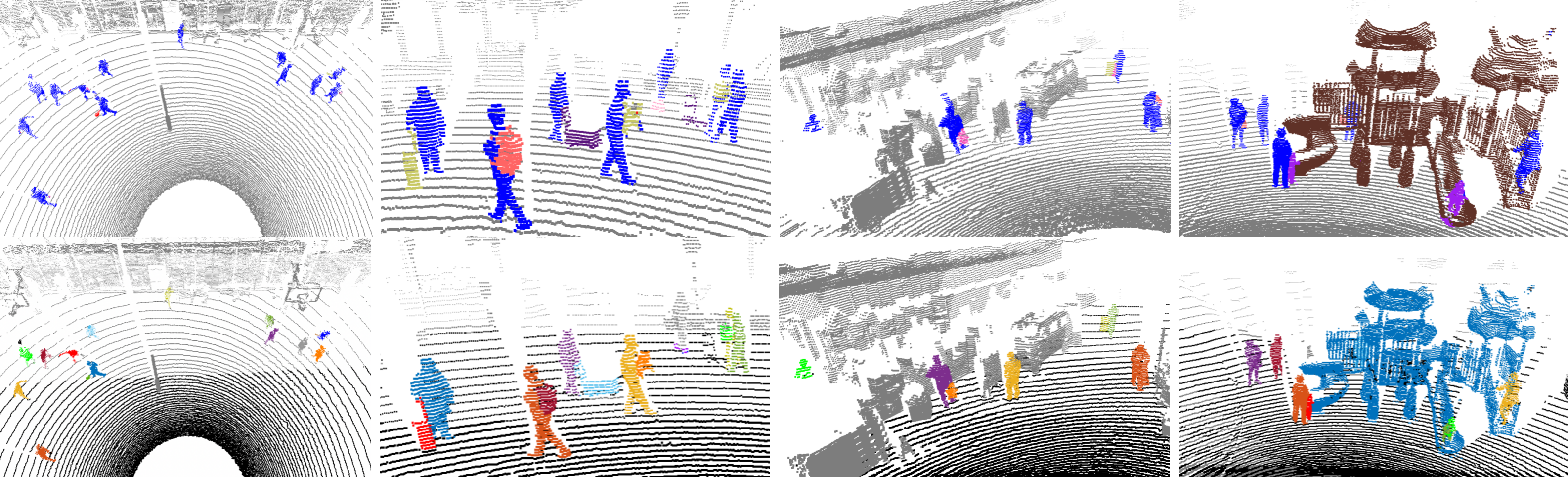}
    \caption{The visualization of semantic (first row) and instance (second row) segmentation results of our method on HuCenLife. }
    \label{fig:vis}
   \vspace{-2ex}
\end{figure}

\subsection{Experiments}
\subsubsection{Baselines and Evaluation Metrics}
Previous 3D instance segmentation works can be divided into LiDAR-based methods and RGB-D-based methods. For the former, we compare with current SOTA method DSNet ~\cite{hong2021lidar} of both voxel-division version and cylinder-division version. For the latter, we select current SOTA approaches DKnet ~\cite{wu20223d} and SoftGroup~\cite{vu2022softgroup} for comparison.

%We present several popular baselines with different modalities for segmentation.
%Previous methods can be divided into LiDAR-based and RGB-D based. We evaluate the result on the LiDAR-based methods Voxel-based DSNet and Cylinder-based DSNet\cite{hong2021lidar}, and the RGB-D based instance segmentation baselines DKnet\cite{wu20223d} and Softgroup\cite{vu2022softgroup}.
%For LiDAR-Camera modality, we apply two fusion strategies (PointPainting and LocalFusion) on our methods. The fusion baselines are provided to facilitate further exploration. PointPainting appends the raw LiDAR point with corresponding RGB color according to calibration matrix. LocalFusion concatenates high-dimensional image feature to the corresponding high dimensional point semantic feature. Our HHOI module has consistently improved the performance on various fusion strategies, validating its generalization ability.
% And we also apply our HHOI on each fusion baseline to validate the generalization ability of our method. 

%\subsubsection{Evaluation Metrics}
We utilize mean IoU (mIoU) to evaluate the quality of the semantic segmentation. For instance segmentation, we report AP50 and AP25 which denote the scores with IoU thresholds of 50\% and 25\%, respectively.

\subsubsection{Results}

\noindent\textbf{Comparison on HuCenLife dataset. } 
We compare the results of our proposed method with baseline methods in Table ~\ref{tab:test}. DSNet does not get satisfactory results, mainly because it focuses on traffic scenarios, while the span of object scale is much larger in human-centric scenarios. SoftGroup is better than outdoor methods because it has a refinement stage for recognizing small objects. Our method performs best due to the use of interaction information.

%With the help of HHOI, the network learns a better human-human interaction representation and the object features are enhance by human-guided feature, and our methods outperforms baselines on contemporary benchmarks.

\noindent\textbf{Comparison on BEHAVE dataset. } 
To further evaluate the generalization capability of human-object interaction scenes, we also conduct experiments for semantic segmentation on BEHAVE~\cite{bhatnagar2022behave} dataset in Table ~\ref{tab:behave}. BEHAVE dataset is a human-object interaction dataset, which is collected in indoor scenarios and provides RGB-D frames and 3D SMPL. To adapt it to our task, we generate the point cloud and segmentation label from RGB-D images and segmented masks. There is only single person with single object per frame and the total number of the object categories is 20. We follow the official protocol of dataset splitting. Our method still outperforms the best baseline method SoftGroup by 2.8\% in mIOU. 
%The results validate the generalizability and the effective ness of our HHOI module.

\noindent\textbf{Sensor-fusion-based 3D segmentation. }
Because our dataset also contains image data, we also provide LiDAR-Camera sensor-fusion baselines based on our method in Table ~\ref{tab:test} to facilitate further research. PointPainting appends the raw LiDAR point with corresponding RGB color according to calibration matrix. LocalFusion concatenates high-dimensional image feature to the corresponding high dimensional point semantic feature. And our HHOI module has consistently improved the performance on various fusion strategies, validating its generalization ability.

%Pointpainting get a comparable result with the LiDAR-only backbone, while LocalFusion perform better with a little margin, which demonstrates the effectiveness to introduce image features in higher dimensions.
%We apply our HHOI on each fusion strategies, our results demonstrate that our module consistently improves the performance on each fusion baseline.

\begin{table}[ht]
  \centering
    \caption{Person-only 3D detection results on HuCenLife. }
    \label{tab:humdetectionresult}
    \setlength{\tabcolsep}{0.8mm}
\begin{tabular}{{l|ccc|l}}
\hline
Methods      & AP(0.25) & AP(0.5) & AP(1.0) & mAP  \\ \hline
CenterPoint\cite{centerpoint}  & 61.8    & 68.7    & 70.3    & 66.9 \\
STCrowd\cite{cong2022stcrowd}      &  61.8    &  71.6   & 73.4   & 68.9 \\
TED\cite{TED}          &  51.0    &  53.3   &  54.1   & 52.8 \\
CenterFormer\cite{centerformer} &   73.0   & 80.1    & 81.4    & 78.2 \\ 
\hline
\end{tabular}
\vspace{-3ex}
\end{table}

\begin{table}[ht]
\caption{Full-category 3D detection results (AP) on HuCenLife. We only select six types of objects for demonstration.}

\label{tab:objectdetection}
\setlength{\tabcolsep}{1.3mm}
\resizebox{\linewidth}{!}{      
\begin{tabular}{l|c|c|c|c|c|c}
\hline
Methods     & motorbike   & box          & cart        & scooter & backpack     & object in hand          \\ \hline
CenterPoint\cite{centerpoint} &13.4  & 17.1          & 20.9             & 43.4 & 4.2          &  8.4             \\ \hline
STCrowd\cite{cong2022stcrowd}     &   5.4        & 14.4 & 25.3  & 48.7          & 4.5 & 13.5 \\ \hline
CenterFormer\cite{centerformer}     & 3.8          & 16.2 & 24.2  & 44.4         & 2.6 & 12.5 \\ \hline
\end{tabular}
}
\vspace{-3ex}
\end{table}
%-------------------------------------------------------------------------
\section{Human-centric 3D Detection}
\label{sec:detect}
\begin{figure*}[t]
    \centering
    \includegraphics[width=2\columnwidth]{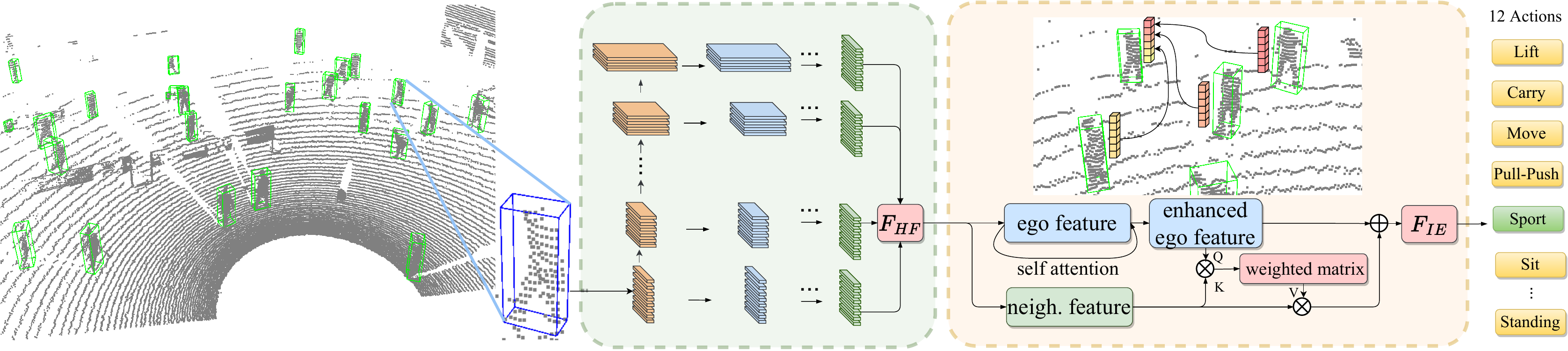}
    \caption{Pipeline of our method for human-centric action recognition. We first utilize 3D detector to obtain a set of bounding boxes of persons. Then, for each person, we extract multi-resolution features and get a hierarchical fusion feature $F_{HF}$. Next, we leverage the relationship with neighbors to enhance the ego-feature and obtain a comprehensive feature $F_{IE}$ for the final action classification.
     %Hierarchical Fusion Feature and $F_{IEE}$ means Interaction Enhanced Ego Feature. The single person's point cloud cropped by bounding box is fed in HPFE to extract multi-scale features. After that, ENFI will explore the interactive relationship between ego feature and neighbour feature.
     }
    \label{fig:MRNetnetwork}
    \vspace{-1ex}
\end{figure*}

LiDAR point cloud-based 3D detection is well-studied in recent years, driven by autonomous driving. It provides critical information of obstacles for the motion planning of robots to guarantee the safety. Specifically, the input for 3D detection is the point cloud $P$ and the output is predicted bounding boxes with 7 dimensions ($x$,$y$,$z$,$w$,$l$,$h$,$r$), consisting of the 3D position in LiDAR coordinate system, the size of bounding box, and the rotation. In this section, we provide benchmarks for the 3D detection task on HuCenLife by evaluating current state-of-the-art methods and give discussion on the research of human-centric 3D detection.

\subsection{Baselines and Evaluation Metrics}
We choose four representative works and test their performance on our dataset. CenterPoint ~\cite{centerpoint} is a popular anchor-free detector and based on it, STCrowd ~\cite{cong2022stcrowd} aims at solving dense crowd scenarios. By means of the transformer mechanism, TED~\cite{TED} and CenterFormer~\cite{centerformer} achieve impressive performance recently. Following ~\cite{cong2022stcrowd,caesar2020nuscenes}, we use Average Precision (AP) with 3D center distance thresholds D = \{0.25, 0.5, 1\} meters as the evaluation metric. Then mean Average Precision (mAP) is obtained by averaging AP.

 % For current SOTA 3D detection. We choose TED\cite{TED} and Centerformer\cite{centerformer}. 
 %Current 3D detection methods TED\cite{TED} and Centerformer\cite{centerformer} utilize transformer strategies, the former extract transformation-equivariant feature for detection and the latter apply a deformable transformer to aggravate feature from  candidate center.
 %We provide the experiments based on above baselines.

%\subsection{Evaluation Metrics}
 %\textbf{Average Precision metric.}Following \cite{nuScenes，STcrowd}, we use Average Precision (AP) with 3D center distance threshold D = {0.25, 0.5, 1} meters. Then mean Average Precision (mAP) is obtained by averaging AP.
%\begin{equation}
%m A P=\frac{1}{|D|} \sum_{d \in D} A P_d
%\end{equation}

\subsection{Results and Discussion}

We conduct experiments on two settings, including \textbf{person-only 3D detection} in Table ~\ref{tab:humdetectionresult} and  \textbf{full-category 3D detection} in Table ~\ref{tab:objectdetection}. These baseline methods are designed for large-scale traffic scenarios, which perform limited on human-centric scenarios, especially for detecting small objects. We conclude with three main challenges for conducting 3D detection in human-centric scenarios. First, people usually have different poses in different actions, such as crouching, sitting, waving, etc., and such diverse body poses cause distinct sizes of bounding box. Second, there are many relatively small objects in scenes, bringing difficulties to balance the accuracy of fine-grained detection and the efficiency of large-scale scene data processing. Third, multi-objects may locate at different heights in the same place, such as in complex scenarios of escalator and slide, leading to larger dimension of feature recognition. Previous methods using BEV feature map will miss details and transformer-based methods have horrible cost. Therefore, there is a lot of room for the 3D detection research in human-centric scenes, while our dataset can offer a good platform for it.

\section{Human-centric Action Recognition}
\label{sec:action}

Previous works for action recognition are based on 2D images or videos and they only need to give one label for one scene. We introduce the 3D action recognition task in large-scale human-centric scenarios, which aims to detect all persons in the scene and provide corresponding action types. 3D action recognition task is significant for fine-grained scene understanding and can benefit the development of intelligent surveillance and collaborative robots. To our knowledge, we are the first to propose the related dataset and solutions for the new task.

%The input of human-centric action recognition is a frame of point cloud $P$ and we propose an end-to-end two-stage model HuCenAction that first detects the human location and then predicts the action category. 
%Human behavior is complex and closely related to the surrounding objects and other people. For one perspective, the size of the objects that interacting with people is various, thus we leverage multi-resolution Hierarchical Point Feature Extraction(HPFE) to capture global features and local features. In addition, considering the interaction between the human and the surrounding neighbours, we design Ego-Neighbour Feature Interaction(ENFI) to get the enhanced human feature and benefit the action classification.
\subsection{Method}
Our 3D action recognition method is in a two-stage manner based on the input of LiDAR point cloud, as shown in Figure ~\ref{fig:MRNetnetwork}. Considering that some human actions are related to adjacent interactive objects, after obtaining individual bounding box by 3D detector, we enlarge the box to crop more points related to the person for the following fine-grained feature extraction. Especially, we leverage a Hierarchical Point Feature Extraction module to pay attention to multi-scale objects and get multi-level features. Moreover, we design an Ego-Neighbour Feature Interaction (ENFI) module to make use of the relationship among the ego-person and neighbors to help forecast social actions. 

\subsubsection{Hierarchical Point Feature Extraction}
%For daily-life scenarios, the objects interacted with people have varying sizes. However, most point cloud classification methods lack the ability to capture local features with diverse receptive fields. 

To capture both global features and local features with dynamically changing receptive fields, we use $R$ parallel branches to extract multi-resolution features.
Serial Set Abstractions~\cite{qi2017pointnet} are applied to process the features of different scales, where each branch undergoes $L$ times with fixed sampling cores and branch-specific sampling range. Finally, these features are up-sampled to the same dimension and fused together with pooling to generate the hierarchical fusion feature $F_{HF}$.

\subsubsection{Ego-Neighbour Feature Interaction}

%Human behavior is also related the behavior of people around him. So we design Ego-Neighbour Feature Interaction(ENFI) to perceive the interaction with neighbors.
Like Figure ~\ref{fig:MRNetnetwork} shows, we first enhance the ego person feature by self-attention and get $F_{ego}$. Then, we select features of $k$ neighbours around the target as $K_{neigh}$ and $V_{neigh}$ and take the ego-feature as queries $Q_{ego}$. The distances from neighbours to the target are used for position encoding. 
%First, the feature of a ego person is enhanced by self-attention, and then the enhanced ego feature $F_{ego}$ will be used as $Q_{ego}$ for following cross attention with neighbours. 
%We select k neighbours feature around the target as $K_{neigh}$ and $V_{neigh}$.
%The distances from neighbours to ego are for position encoding.
We apply cross-attention to extract the ego-neighbour interaction information and gain the final interaction enhanced ego feature by $F_{IE} = F_{ego} \bigoplus \text{CrossAttention}(Q_{ego},K_{neigh},V_{neigh})$, where $\bigoplus$ denotes concatenation. In this way, we model the relationships of a group to benefit the social action recognition.

\begin{table}[ht]
\caption{Comparison results of action recognition on HuCenLife. All methods are based on the same 3D detector for fair evaluation.}
\label{tab:actioncompare}
\begin{tabular}{l|ccc}

\hline
\multicolumn{1}{l|}{Methods} & mAP  & mRecall & mPrecision \\ \hline
Baseline & 7.3  & 14.6 & 19.9\\\hline
+ ViT\cite{dosovitskiy2020image}                          & 9.4  & 23.1 & 19.9  \\ \hline

+ PVT\cite{zhang2022pvt}                          & 13.2  & 30.5 & 19.8  \\
+ PointNet\cite{qi2017pointnet}                     & 8.4  & 26.3 & 15.5  \\
+ PointNet++\cite{qi2017pointnet++}              & 15.6  & 34.2 & 22.7  \\
+ PointMLP\cite{ma2022rethinking}                     & 11.3  & 28.0 & 19.4  \\
+ PointNeXt\cite{qian2022pointnext}                    & 15.0  & 33.0 & 21.2  \\ \hline

Ours                  & \textbf{21.0} & \textbf{40.0} & \textbf{26.9}  \\ 
Ours(w/o ENFI)         & 15.4  & 37.1 & 24.7  \\
\hline

\end{tabular}
\vspace{-3ex}
\end{table}

\subsection{Experiments}

\subsubsection{Baselines and Evaluation Metrics}
We take pre-trained CenterPoint as the 3D Detector for all the experiments for fair comparison in this section. Because no existing methods can be directly used for solving the new 3D action recognition task. As Table ~\ref{tab:actioncompare} shows, we provide benchmarks and comparisons from four aspects. The first is to directly adapt the 3D detector to predict multi-class persons with different action labels, which is the ``Baseline'' in Table~\ref{tab:actioncompare}. The second is to add a feature extractor for cropped individual point cloud for the second-stage action classification, and we tried several popular point-feature extractors, including PVT, PointNet, PointNet++, PointMLP, and PointNext. In particular, to verify the performance of input modalities, we also use ViT to extract image features for image-based action recognition by projecting the 3D bounding box to calibrated images. At last, we provide the results of our solution with ablation for ENFI module. 

%to crop the single point cloud for it performs best in human detection , and present several point-based classification networks: PointNet, PointNet++, PointMLP and PointNext. We also explore Voxel-based methods such as Point Voxel Transformer.

%There are some ablation studies on modalities comparation. We use 2D bounding box projected from 3D bounding box to get single person's image as the input of Vision Transformer. In addition, we explore the performance gain of ENFI for action recognition.
%\subsubsection{Evaluation Metrics}
% Human-centric Action Recognition is a point cloud classification task. We divide the actions of all people in the data set into 17 categories, and predict an action tag for each person.

 %\textbf{Average Precision metric}. 
 %Following \cite{nuScenes} and \cite{STcrowd}, we use Average Precision (AP) with 3D center distance threshold $D =  \{0.25, 0.5, 1\}$ meters for each action class $C$. 
We use the mean Average Precision (mAP) obtained by averaging AP through thresholds $D =  \{0.25, 0.5, 1\}$ and classes to evaluate the performance.
$$
\mathrm{mAP}=\frac{1}{|\mathbb{C}||\mathbb{D}|} \sum_{c \in \mathbb{C}} \sum_{d \in \mathbb{D}} \mathrm{AP}_{c, d}
$$
where $|\mathbb{C}|$ is the number of action category.
In addition, we also utilize Mean Recall (mRecall) and Mean Precision (mPrecision) by averaging recall and precision through thresholds and classes.

\subsubsection{Results and Discussion}
We show the overall performance in Table ~\ref{tab:actioncompare}, and detailed evaluation values of all categories of actions and visualization results are in the supplementary material. It can be seen from the results that our method outperforms others with an obvious margin, mainly due to the multi-level feature extraction and multi-person interaction modeling, which are more suitable for understanding human-centric complex scenarios. However, our method has its own limitations and there are several potential improvement directions. First, current two-stage framework strongly relies on the detector performance and the one-stage method for action recognition in large-scale scenes is worth exploring. Moreover, human action is time-dependent and how to extract valuable temporal information in consecutive data to eliminate the ambiguity of actions is also promising.

\section{More Tasks on HuCenLife}
\label{sec:moretask}
In this paper, we provide benchmarks on HuCenLife for three main tasks, including 3D segmentation, 3D detection, and action recognition in human-centric scenarios. However, benefiting from the rich annotations in HuCenLife dataset, there are many other tasks deserving explored. %We discuss several meaningful topics in the following.

\subsection{Human-Object Interaction Detection}
Recently, the task of Human-Object Interaction (HOI) detection~\cite{zhang2022exploring,yuan2022detecting} attracts more and more attention, which targets for detecting the person and the interacted object and meanwhile classifying the interaction category. Current studies and datasets are limited to the interaction between single person and single object in one scene and they are all based on the image modality. 3D HOI tasks in large-scale free environments with multiple persons and multiple objects can be formulated and evaluated on HuCenLife.

\subsection{Tracking and Trajectory Prediction}
HuCenLife contains sequential frames of data with the tracking ID annotation for all instances, which can facilitate the time-related tasks, such as 3D tracking~\cite{pttp,zheng2022beyond} and trajectory prediction~\cite{ma2019trafficpredict,deo2022multimodal}. It is challenging for these tasks due to the occlusions in crowded scenes, but it is significant to study consecutive behaviors and interactions in real world to provide valuable guidance for robots.

\subsection{3D Scene Generation}
With the success of Diffusion model~\cite{ho2020denoising} in image generation, many works try to achieve high-quality 3D data generation for single objects~\cite{luo2021diffusion} or scenes~\cite{zyrianov2022learning}. HuCenLife provides rich material for daily-life scenarios, and it is interesting to generate more human-centric scene data with semantic information to facilitate learning-based methods.

\subsection{Multi-modal Feature Fusion}
Apart from point cloud, HuCenLife also provides corresponding images. The complementary information of multi-modal features will definitely benefit all tasks mentioned above, which deserves in-depth research.

%-------------------------------------------------------------------------

\section{Conclusion}
\label{sec:conclusion}
We fully discuss the challenges, significance, and potential research directions of 3D human-centric scene understanding in this paper. Specifically, we propose the first related large-scale dataset with rich fine-grained annotations, which can facilitate the research for many 3D tasks and has the potential to boost the development of assistive robots, surveillance, etc. Moreover, we provide benchmarks for various tasks and propose novel methods for human-centric 3D segmentation and human-centric action recognition to facilitate further research. 
%-------------------------------------------------------------------------
%\section{Acknowledgements}
%This work was supported by NSFC (No.62206173), Natural Science Foundation of Shanghai (No.22dz1201900), MoE Key Laboratory of Intelligent Perception and Human-Machine Collaboration (ShanghaiTech University), Shanghai Frontiers Science Center of Human-centered Artificial Intelligence (ShangHAI), Shanghai Engineering Research Center of Intelligent Vision and Imaging.
{\small
\bibliographystyle{ieee_fullname}
\bibliography{egbib}
}

%%%%%%%%% APPENDIX
\appendix
\label{sec:appendix}
\clearpage
% \documentclass[10pt,twocolumn,letterpaper]{article}

% \usepackage{iccv}
% \usepackage{times}
% \usepackage{epsfig}
% \usepackage{graphicx}
% \usepackage{amsmath}
% \usepackage{amssymb}

% \usepackage{color}
% \usepackage{times}
% \usepackage{overpic}
% \usepackage{bm}
% \usepackage{tabu}
% \usepackage{bbding}
% \usepackage{multicol}
% \usepackage{multirow}
% \usepackage[table]{xcolor}

% % Include other packages here, before hyperref.

% % If you comment hyperref and then uncomment it, you should delete
% % egpaper.aux before re-running latex.  (Or just hit 'q' on the first latex
% % run, let it finish, and you should be clear).
% \usepackage[pagebackref=true,breaklinks=true,letterpaper=true,colorlinks,bookmarks=false]{hyperref}

% % \iccvfinalcopy % *** Uncomment this line for the final submission

% \def\iccvPaperID{3694} % *** Enter the ICCV Paper ID here
% \def\httilde{\mbox{\tt\raisebox{-.5ex}{\symbol{126}}}}

% % Pages are numbered in submission mode, and unnumbered in camera-ready
% \ificcvfinal\pagestyle{empty}\fi

\begin{appendix}

% %%%%%%%%% TITLE
% \title{Supplement of Human-centric Scene Understanding for 3D Large-scale Scenarios}

% \and

% \maketitle
% % Remove page # from the first page of camera-ready.

%%%%%%%%% ABSTRACT
\section{Implement details}

\subsection{Human-centric Instance Segmentation}
In HHOI module, the threshold  $\tau$ for sampling high confidence features is set to 0.8 and the number of sampled points $M=256$.
In Point-wise Prediction and Refinement process,
the loss can be formulated as following: 
$\mathcal{L} = \mathcal{L}_{\text {semantic}}+\mathcal{L}_{\text {offset}}+\mathcal{L}_{\text {class}}+\mathcal{L}_{\text {mask}}+\mathcal{L}_{\text{mask score}}.$
$$
L_{\text {semantic }}=\frac{1}{N} \sum_{i=1}^N \operatorname{CE}\left(\boldsymbol{s}_i, s_i^*\right),$$
$$
L_{\text {offset }}=\frac{1}{\sum_{i=1}^N \mathbb{I}_{\left\{\boldsymbol{p}_i\right\}}} \sum_{i=1}^N \mathbb{I}_{\left\{\boldsymbol{p}_i\right\}}\left\|\boldsymbol{o}_i-\boldsymbol{o}_i^*\right\|_1,
$$
$$
L_{\text {class }}=\frac{1}{K} \sum_{k=1}^K \mathrm{CE}\left(\boldsymbol{c}_k, c_k^*\right), 
$$
$$
L_{\text {mask }}=\frac{1}{\sum_{k=1}^K \mathbb{I}_{\left\{\boldsymbol{m}_k\right\}}} \sum_{k=1}^K \mathbb{I}_{\left\{\boldsymbol{m}_k\right\}} \mathrm{BCE}\left(\boldsymbol{m}_k, \boldsymbol{m}_k^*\right), 
$$
$$
\mathcal{L}_{\text{mask score}}=\frac{1}{\sum_{k=1}^{N_{g t}} \mathbb{I}_{\left\{iou_k\right\}}} \sum_{k=1}^{N_{g t}} \mathbb{I}_{\left\{iou_k\right\}}\left\|iou_k-iou_k^*\right\|_2
$$
where $*$ denotes the ground truth.

\subsection{Human-centric Action Recognition}
% \begin{figure*}[t]
%    \centering
%    \includegraphics[width=2\columnwidth]{pic/action_recognition_visulization.png}\caption{Action Recognition visulization. The left figure shows the 3D human bounding box and corresponding actions predicted by our method. The right figure is the ground truth.}
%    \label{fig:vis}
% \end{figure*}
The input for action recognition is frames of large scene point cloud $P\in R^{N\times 4}$ with the 3D location and reflection intensity (x, y, z, r). We extend the length and width of bounding box obtained from human detector by $\Delta h$ and $\Delta w$ respectively, where $\Delta h$ and $\Delta w$ are both set to 0.2 meters. After cropping point clouds with bounding boxes, we use clustering algorithm to find k(k=3) nearest neighbors of the ego point cloud with their relative distances. Next, the point cloud of every single person will be normalized, and sampled by farthest point sample algorithm to n points(n=512). The features of k neighbours and ego will be extracted by HPFE simultaneously to get features of dimension $(k+1) \times c$, which will be input to ENFI afterwards.

In HPFE, we use set abstractions(SA) to down-sample R times on origin point clouds to fork R branches with different resolutions. R is set to 5 by default.
$$P_i \in R^{(n/2^r) \times (32*2^r)} r\in [1,...,R],i\in [1,...,L]$$
where $P_i$ is the feature dimension of R branches. Then we use different sampling radius for the R resolution branches, which are $0.05*(r+1),r\in [1,...,R]$, so that the receptive field of SA will expand with the improving of resolution. After that, we apply equal sampling for L times(L is set to 2) for all branches simultaneously. Finally, we down sample the features of the low-resolution channels to get five features of the same size, which will be fused together to get hierarchical fusion feature.

\section{Dataset details}
\begin{figure}[t]
    \centering
    \includegraphics[width=1\columnwidth]{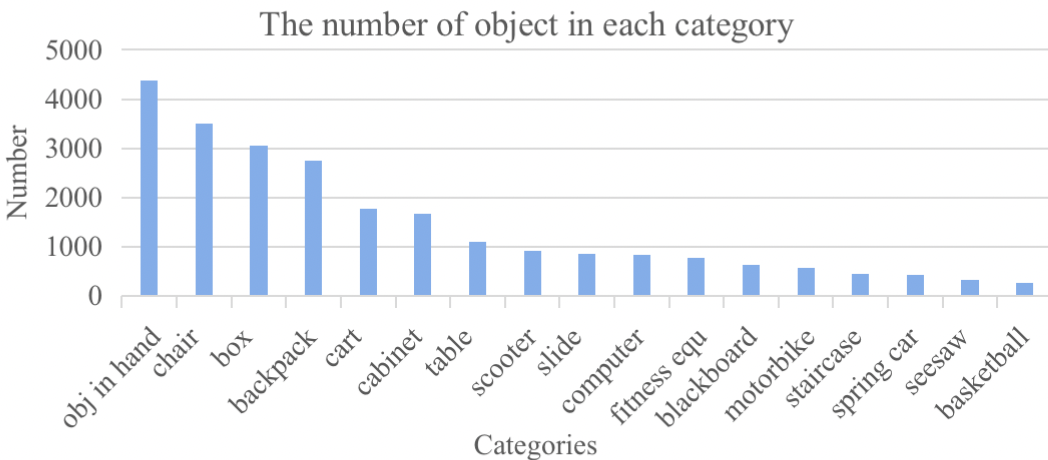}
    \caption{The number of the object for each object.}
    \label{fig:segcount}
    \vspace{-2ex}
\end{figure}

\subsection{Object category for segmentation and detection}
We merge several categories which have low frequency of occurrence and similar geometry shapes in our dataset into a new class, and we also drop some category which only appear in training or testing set with low frequency. The merging list is shown in Table ~\ref{tab:merge}. The categories of objects after merging is 17 and the number of objects in each category is illustrated in Figure. \ref{fig:segcount}.
% \usepackage{multirow}

\begin{table}[ht]
\centering
\caption{Object merging list. We merge the categories on the left into the category on the right.}
    \label{tab:merge}
    \setlength{\tabcolsep}{0.8mm}
\begin{tabular}{c|c}
\hline
banner,plank,paper,door,dog,megaphone,guitar  & \multicolumn{1}{c}{\multirow{2}{*}{other}}       \\
toy car,merry go round,car,tricycle,umbrella, & \multicolumn{1}{c}{}                             \\\hline
printer,podium                                &
cabinet                                          \\\hline
bicycle                                       & motorbike                                        \\\hline
{(}two-wheeled{)} {(} self-{)}balancing car   & scooter                                          \\\hline
flat car,stroller,perambulator &cart                                        \\\hline
rockery                                       & slide                                            \\\hline
stool                                         & chair                                            \\\hline
suitcase                                      & box                                              \\\hline
eraser,phone,cup,food,cellphone,red flag,     & \multicolumn{1}{c}{\multirow{4}{*}{obj in hand}} \\
cap,camera,sponge,projector,balloon,          & \multicolumn{1}{c}{}                             \\
plush toy,toy wings,clothes,flower,           & \multicolumn{1}{c}{}                             \\
badminton rocket,handbag,plastic bag,         &
\\\hline\multicolumn{1}{c}{}                            
\end{tabular}
\end{table}

\subsection{Action category for  recognition and detection}
\begin{figure}[t]
    \centering
    \includegraphics[width=0.93\columnwidth]{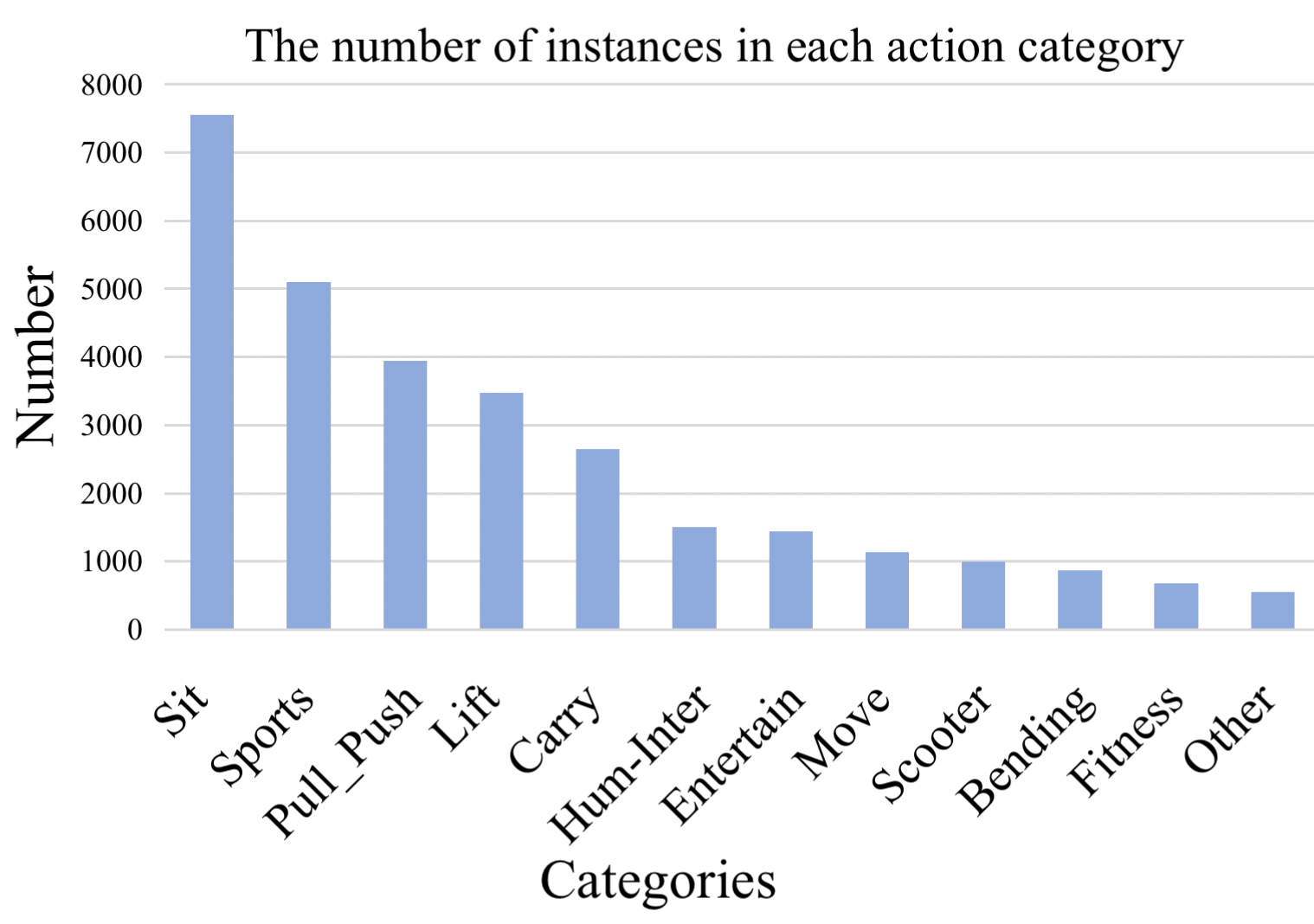}
    \caption{The number of instance in each merged action category.}
    \label{fig:action_count}
    \vspace{-2ex}
\end{figure}

 It is common for a person to perform multiple actions simultaneously. To prioritize these actions, we assign each action to a numerical priority value. We then merge these prioritized actions into 12 categories based on their similarity and frequency of occurrence. Actions with low frequency are dropped to ensure a manageable number of categories. To illustrate this process, we provide a merging Table ~\ref{tab:action_merge} that maps each prioritized action to its corresponding category. The number of each action after the merging process is shown in Figure.~\ref{fig:action_count}.

% \usepackage{multirow}
\begin{table}[ht]
\caption{Detailed action priority and merge information.}
\centering
\label{tab:action_merge}
\resizebox{0.85\linewidth}{!}{   
\begin{tabular}{l|c|l}
\hline
Merged   Action                      & priority            & Original Action                    \\ \hline
\multirow{7}{*}{Lift}                & \multirow{7}{*}{0}  & taking clothes                     \\ \cline{3-3} 
                                     &                     & lifting a   plastic bag            \\ \cline{3-3} 
                                     &                     & lifting a bag                      \\ \cline{3-3} 
                                     &                     & taking   things/exchanging items   \\ \cline{3-3} 
                                     &                     & lifting   things                   \\ \cline{3-3} 
                                     &                     & lifting something                  \\ \cline{3-3} 
                                     &                     & moving planks                      \\ \hline
\multirow{3}{*}{Carry}               & \multirow{3}{*}{1}  & carrying other things              \\ \cline{3-3} 
                                     &                     & carrying a bag                     \\ \cline{3-3} 
                                     &                     & carrying bags                      \\ \hline
Move                                 & 2                   & moving boxes                       \\ \hline
\multirow{10}{*}{Pull\_Push}          & \multirow{10}{*}{3} & pulling a suitcase                 \\ \cline{3-3} 
                                     &                     & pulling a chair                    \\ \cline{3-3} 
                                     &                     & pulling a flatcar                  \\ \cline{3-3} 
                                     &                     & pushing a cart                     \\ \cline{3-3} 
                                     &                     & pushing a stroller                 \\ \cline{3-3} 
                                     &                     & pushing a flatcar                  \\ \cline{3-3} 
                                     &                     & pushing a table                    \\ \cline{3-3} 
                                     &                     & holding a spring car               \\ \cline{3-3} 
                                     &                     & pushing something                  \\ \cline{3-3} 
                                     &                     & pushing something                  \\ \hline
\multirow{14}{*}{Sit}                & \multirow{5}{*}{4}  & riding a bicycle                   \\ \cline{3-3} 
                                     &                     & riding an electric bicycle         \\ \cline{3-3} 
                                     &                     & riding a tricycle                   \\ \cline{3-3} 
                                     &                     & riding on the carousel             \\ \cline{3-3} 
                                     &                     & sitting in a spring car            \\ \cline{2-3} 
                                     & \multirow{9}{*}{13} & crouching                          \\ \cline{3-3} 
                                     &                     & sitting on the ground              \\ \cline{3-3} 
                                     &                     & crouching or sitting on the ground \\ \cline{3-3} 
                                     &                     & sitting on the ground              \\ \cline{3-3} 
                                     &                     & sitting                            \\ \cline{3-3} 
                                     &                     & sitting on a trunk                 \\ \cline{3-3} 
                                     &                     & sitting in a chair                 \\ \cline{3-3} 
                                     &                     & sitting on the stool               \\ \cline{3-3} 
                                     &                     & squatting                          \\ \hline
\multirow{5}{*}{Scooter-BalanceBike} & \multirow{5}{*}{5}  & riding a two-wheel balance car     \\ \cline{3-3} 
                                     &                     & riding a balance car               \\ \cline{3-3} 
                                     &                     & riding an electric skateboard      \\ \cline{3-3} 
                                     &                     & riding a skateboard                \\ \cline{3-3} 
                                     &                     & standing on a trolley              \\ \hline
\multirow{8}{*}{Hum-Inter}           & \multirow{8}{*}{6}  & hugging                            \\ \cline{3-3} 
                                     &                     & pulling a baby                     \\ \cline{3-3} 
                                     &                     & being hold by someone else         \\ \cline{3-3} 
                                     &                     & taking a baby                      \\ \cline{3-3} 
                                     &                     & holding the baby                   \\ \cline{3-3} 
                                     &                     & Being held by someone else         \\ \cline{3-3} 
                                     &                     & carrying a baby                    \\ \cline{3-3} 
                                     &                     & being carry by someone else        \\ \hline
\multirow{3}{*}{Fitness}             & \multirow{3}{*}{7}  & fitness with a twister             \\ \cline{3-3} 
                                     &                     & fitness with a elliptical trainer  \\ \cline{3-3} 
                                     &                     & fitness with a stepper             \\ \hline
\multirow{6}{*}{Entertain}           & \multirow{6}{*}{8}  & climbing the swing                 \\ \cline{3-3} 
                                     &                     & climbing slide                     \\ \cline{3-3} 
                                     &                     & holding the slide                  \\ \cline{3-3} 
                                     &                     & sliding                            \\ \cline{3-3} 
                                     &                     & playing seesaw                     \\ \cline{3-3} 
                                     &                     & sitting in a cavern                \\ \hline
\multirow{2}{*}{Sports}              & 9                   & playing basketball                 \\ \cline{2-3} 
                                     & 10                  & playing badminton                  \\ \hline
\multirow{5}{*}{Standing}            & 11                  & taking the escalator               \\ \cline{2-3} 
                                     & 14                  & running                            \\ \cline{2-3} 
                                     & \multirow{3}{*}{15} & walking                            \\ \cline{3-3} 
                                     &                     & standing                           \\ \cline{3-3} 
                                     &                     & leaning                            \\ \hline
Bending$\_$Over                         & 12                  & bending over                       \\ \hline
\multirow{8}{*}{Other}               & \multirow{8}{*}{16} & cabinet interaction                \\ \cline{3-3} 
                                     &                     & standing on the stool              \\ \cline{3-3} 
                                     &                     & getting in the car                 \\ \cline{3-3} 
                                     &                     & getting out of the car             \\ \cline{3-3} 
                                     &                     & driving a toy car                  \\ \cline{3-3} 
                                     &                     & lying                              \\ \cline{3-3} 
                                     &                     & writing on the blackboard          \\ \cline{3-3} 
                                     &                     & …                                  \\ \hline
\end{tabular}
}
\end{table}

\section{More Experiment}
\begin{figure}[ht]

    \centering

\includegraphics[width=1\columnwidth]{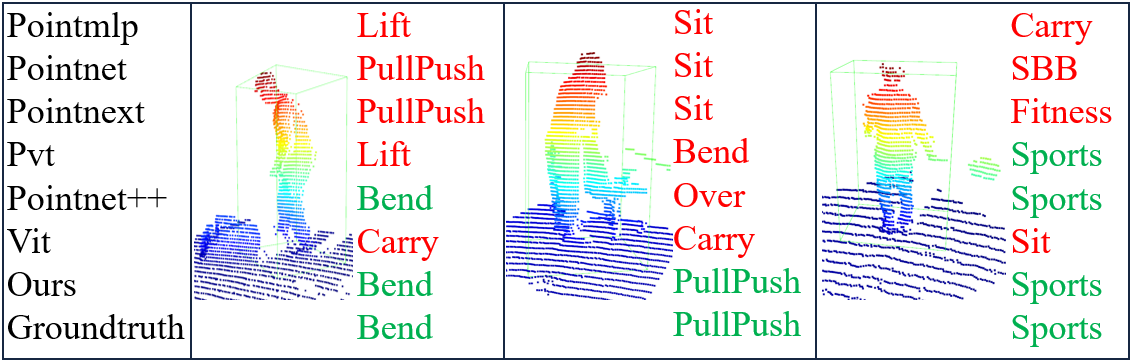}
   % \caption{Visualization for action recognition. SBB denotes the class Scooter-BalanceBike, the words in green are the correct prediction and the red are the wrong predictions.}
    \label{fig:vis-action}
\end{figure}

We take pre-trained CenterPoint as the 3D Detector and add a feature extractor for cropped individual point cloud for the second-stage action recognition comparison, the detailed comparison result is shown in Table ~\ref{tab:action_exp1}. Our method outperforms others in most of categories. The comparison result which uses 3D bounding boxes from ground truth is shown in Table ~\ref{tab:action_abcde_ground}. We further provide
action visualization in Figure ~\ref{fig:vis-action}.
% \begin{wrapfigure}{r}{3.5cm}

\begin{table*}[ht]
\caption{Detailed comparison results of action recognition on HuCenLife. All methods are based on the same 3D detector (centerpoint) for fair evaluation.}
\label{tab:action_exp1}

\setlength{\tabcolsep}{1.3mm}
\resizebox{\linewidth}{!}{     
\begin{tabular}{c|c|c|c|c|c|c|c|c|c|c|c|c|c|c|c}
\hline
Method           & Lift         & Carry         & Move         & Pull\_Push     & Sit           & Scooter-BalanceBike & Hum-Inter    & Fitness       & Entertain     & Sports        & Bend-Over    & Standing      & mAP         & mRecall     & mPrec         \\ \hline
Baseline         & 0.5          & 1.6           & 0.2           & 13.8          & 2.2           & 21.8                & 0            & 0             & 2.4           & 6.9           & 0.1          & \textbf{38.3} & 7.3         & 14.6        & 19.9          \\ \hline
ViT              & 4.1          & 1.6           & 5.1           & 8.2           & 0.6           & 4.7                 & 0.1          & \textbf{27.3} & 6             & \textbf{46.6} & 0.1          & 8.3           & 9.4         & 23.1        & 19.9          \\ \hline
PVT              & 1.4          & 10.5          & 8.9           & 21            & 16.8          & 56.8                & 5.9          & 1.7           & 1             & 25.1          & 4.3          & 5.2           & 13.2        & 30.5        & 19.8          \\ \hline
PointNet         & 1.6          & 3.1           & 4.6           & 20.1          & 24.4          & 22.3                & 0.7          & 0.6           & 0.6           & 17.1          & 1.5          & 4.2           & 8.4         & 26.3        & 15.5          \\ \hline
PointNet++       & 3.6          & 25.3          & 10.6          & 21            & 25.5          & 51                  & 3.5          & 2.7           & 3.3           & 30.3          & 4.1          & 6.5           & 15.6        & 34.2        & 22.7          \\ \hline
PointMLP         & 2.9          & 4.1           & 7.6           & 24.6          & 23.6          & 34.4                & 2.8          & 1.8           & 2.7           & 25.4          & 1.6          & 3.9           & 11.3        & 28          & 19.4          \\ \hline
PointNeXt        & 2            & 13.3          & 15.2          & 26.1          & 12.8          & 61.1                & 5.4          & 4.7           & 1.7           & 26.6          & 3.2          & 8.4           & 15          & 33          & 21.2          \\ \hline
Ours             & 5            & \textbf{26.5} & \textbf{20.1} & \textbf{35.8} & \textbf{26.5} & \textbf{68.5}       & 6.8          & 6.2           & \textbf{11.2} & 30.4          & 4.5          & 10.8          & \textbf{21} & \textbf{40} & \textbf{26.9} \\ \hline
Ours(w/o   ENFI) & \textbf{6.1} & 16.7          & 16.8          & 31            & 18.4          & 55.8                & \textbf{7.8} & 3.9           & 1.3           & 11.7          & \textbf{4.6} & 10.9          & 15.4        & 37.1        & 24.7          \\ \hline
\end{tabular}
}
\end{table*}

\begin{table*}[ht]
\caption{Detailed comparison results of action recognition on HuCenLife. All methods are based on the ground truth bounding boxes. mAcc stands for mean accuracy.}
\label{tab:action_abcde_ground}

\setlength{\tabcolsep}{1.3mm}
\resizebox{\linewidth}{!}{     
\begin{tabular}{c|c|c|c|c|c|c|c|c|c|c|c|c|c}
\hline
Method         & Lift & Carry & Move & Pull\_Push & Sit  & Scooter-BalanceBike & Hum-Inter & Fitness & Entertain & Sports & Bend-Over & Standing & mAcc \\ \hline
ViT            & 9.1  & 10.7  & 26.2 & 36.3       & 25.3 & 15.2               & 1.9      & 51.6    & 50.9      & 65.5   & 13.5      & 16.0     & 26.9 \\ \hline
PVT            & 4.5  & 42.8  & 31.2 & 35.6       & 40.0 & 74.7               & 7.2      & 36.4    & 0.4       & 16.2   & 54.4      & 31.6     & 31.3 \\ \hline
PointNet       & 7.8  & 29.1  & 32.8 & 33.2       & 47.2 & 53.1               & 7.5      & 46.9    & 19.1      & 20.1   & 57.4      & 20.9     & 31.3 \\ \hline
PointNet++     & 11.1 & 41.1  & 37.7 & 23.5       & 66.7 & 80.3               & 15.5     & 39.3    & 55.4      & 11.4   & 30.3      & 8.6      & 35.1 \\ \hline
PointMLP       & 25.6 & 46.4  & 35.4 & 57.2       & 55.2 & 79.7               & 4.9      & 54.5    & 27.8      & 15.3   & 29.1      & 32.8     & 38.7 \\ \hline
PointNext      & 11.8 & 46.7  & 24.0 & 49.4       & 50.1 & 76.1               & 21.6     & 46.9    & 36.5      & 10.2   & 36.2      & 53.0     & 38.5 \\ \hline
Ours           & 19.8 & 38.9  & 30.0 & 59.8       & 62.5 & 86.6               & 62.5     & 61.8    & 32.4      & 18.2   & 35.0      & 24.8     & \textbf{44.4} \\ \hline
Ours(w/o ENFI) & 18.9 & 49.5  & 47.6 & 57.2       & 53.3 & 83.1               & 28.8     & 31.5    & 31.2      & 19.2   & 53.6      & 33.8     & 42.3 \\ \hline
               
\end{tabular}
}
\end{table*}

%-------------------------------------------------------------------------

% {\small
% \bibliographystyle{ieee_fullname}
% \bibliography{egbib}
% }

\end{appendix}

\end{document}